\documentclass[11pt,a4paper]{article}
\usepackage[hyperref]{emnlp-ijcnlp-2019}

\usepackage[T2A,T1]{fontenc}
\usepackage[utf8]{inputenc} 
\usepackage{times}
\usepackage{latexsym}

\usepackage{amsmath}

\usepackage{url}

\usepackage{multirow}
\usepackage{graphicx}

\aclfinalcopy 

\title{Fill in the Blanks: Imputing Missing Sentences for Larger-Context Neural Machine Translation}

\author{S\'ebastien Jean \\
  Google AI / NYU \\
  {\tt sebastien@cs.nyu.edu} \\\And
  Ankur Bapna \\
  Google AI \\
  {\tt ankurbpn@google.com} \\\And
  Orhan Firat \\
  Google AI \\
  {\tt orhanf@google.com} \\
  }

\date{}

\begin{document}
\maketitle

\begin{abstract}
    Most neural machine translation systems still translate sentences in isolation. To make further progress, a promising
    line of research additionally considers the surrounding context in order to provide the model potentially missing source-side information, as well as to maintain a coherent output. One difficulty in training such larger-context (i.e. document-level) machine translation systems is that context may be missing from many parallel examples. To circumvent this issue, two-stage approaches, in which sentence-level translations are post-edited in context, have recently been proposed. In this paper, we instead consider the viability of filling in the missing context. In particular, we consider three distinct approaches to generate the missing context: using random contexts, applying a copy heuristic or generating it with a language model. In particular, the copy heuristic significantly helps with lexical coherence, while using completely random contexts hurts performance on many long-distance linguistic phenomena. We also validate the usefulness of tagged back-translation. In addition to improving BLEU scores as expected, using back-translated data helps larger-context machine translation systems to better capture long-range phenomena.

\end{abstract}

\section{Introduction}

Neural machine translation~\citep{sutskever2014sequence, bahdanau2014neural} has gained considerable popularity, attracting significant research interest and being widely adopted in industry. While results are impressive, sometimes arguably rivaling human quality~\citep{hassan2018achieving}, there is room for further progress as most neural translation systems are trained to translate sentences in isolation~\citep{laubli2018has}. This may lead to undesirable behaviour when translating longer text segments, such as paragraphs or documents. For example, the output may contain incorrect gender markers or generally be inconsistent~\cite{popescu2019context}.

Given these limitations, there has been growing interest in document-level neural machine translation, such as in the En-De and En-Cs shared tasks at WMT'19~\cite{barrault-etal-2019-findings}. In this work, we consider a scenario where we are provided a combination of document-level and sentence-level only parallel data. Previous work~\cite{voita-etal:2019b:ACL} has tackled this problem with a two-stage approach, initially generating sentences separately and then correcting the output in context. Here, we instead explore the viability of generating the missing context.

We consider multiple techniques to complete the missing data. In all cases, we do not modify the examples where context is already provided. As baselines, we may leave the context-less examples intact, or simply add random sentence pairs to such examples~\citep{junczysdowmunt:2019:WMT}. We also present a simple partial copy heuristic, where training examples are constructed by combining multiple copies of a source-target pair with random ones.
Alternatively, we may sample target-side context with a language model and obtain the corresponding source context through back-translation. Finally, we also consider retrieving similar sentence pairs within the training data to use as context.

By evaluating these context-aware models on multiple challenge sets~\cite{voita-etal:2019b:ACL} targeting specific longe-range phenomena, we observe that the choice of context completion technique clearly impacts performance. In particular, the partial copy heuristic leads to significantly better lexical coherence. In contrast, creating examples where all sentences are unrelated is harmful on all challenge sets, even if this negative impact is not necessarily visible from BLEU scores.

Given the importance of appropriate data to capture long-range phenomena, we also validate the effectiveness of (tagged) back-translation~\cite{sennrich2016improving, caswell:2019:WMT, junczysdowmunt:2019:WMT} for document-level machine translation. Adding back-translated data is clearly helpful across all specific linguistic phenomena we evaluate, although the initial context completion technique still visibly impacts performance.  Back-translation also increases the general translation quality as measured by BLEU, although we see similar gains for sentence-level models.

The contributions of this paper are three folds:

\begin{itemize}
    \item {In a scenario where only some of the parallel data has document-level context, we establish the viability of completing the missing data.}
    \item{We present multiple approaches to fill in the missing context and evaluate their impact on multiple linguistic phenomena.}
    \item{We validate the effectiveness of (tagged) back-translation for document-level machine translation.}
    
\end{itemize}

\section{Background}

\subsection{Neural machine translation}

A typical neural machine translation system~\citep{sutskever2014sequence, bahdanau2014neural} is a neural network mapping an input sequence $X$ to its translation $Y$. Words (or subword tokens) within $X$ are mapped to a continuous space with word embeddings. The embeddings are consumed by an encoder network which produces one or multiple source representations $\text{Enc}(X; \theta_{Enc})$. Using these source representations, a decoder network predicts the output, often auto-regressively $p(y_t) = \text{Dec}(y_{<t}, \text{Enc}(X); \theta_{Dec})$, although other factorizations may be employed~\cite{mansimov2019generalized}. Both the encoder and decoder may be implemented with recurrent, convolutional or self-attentive networks, among others~\citep{bahdanau2014neural, gehring2017convolutional, vaswani2017attention}. The encoder and decoder are generally separate networks, although they can alternatively be merged~\citep{he2018layer}.

Given a set of parallel sentence pairs $\{(X_1,Y_1), ..., (X_n,Y_n)\}$ of size $N$, neural machine translation systems are often optimized by minimizing the negative log-likelihood: 

\begin{equation*}
    \frac{1}{N}\sum_{n=1}^N \log p_\theta(Y_n|X_n).
\end{equation*}

To generate translations, given the intractability of $\text{arg}\max_{Y} p(Y|X)$, at least under the auto-regressive framework, it is common to use approximate search algorithms, i.e. beam search, which prunes the output of a breadth-first search at every time step.

\subsection{Document-level neural machine translation}

Sentence-level systems are unable to model some long-distance phenomena correctly. For example, in some languages such as Spanish, pronouns may be dropped. As such, without additional context, an English translation would have to guess the correct gender. Moreover, politeness markers, named entities and other words may also be translated inconsistently across sentences.

To correct such weaknesses, models may use additional context information, for example neighbouring sentences within a document. Document-level machine translation approaches can currently be broadly classified into three categories. First, the additional context may simply be concatenated to the sentence to translate~\citep{tiedemann2017neural, junczysdowmunt:2019:WMT}, which allows using the same models as for sentence-level translation. The additional target-side context may either be predicted or not by such models. Alternatively, contextual information may be used as a separate input to the model, in which case the model architecture may be adapted in consequence~\citep{jean2017does, wang2017exploiting, voita2018context, zhang2018improving, miculicich2018document, maruf2017document, tu2018learning}. Finally, sentence-by-sentence translations may be post-edited to integrate contextual information~\citep{voita-etal:2019b:ACL, voita-etal-2019-context}.

\subsection{Evaluating contextual translation systems}

The most commonly used metric to evaluate machine translation output is BLEU~\citep{papineni2002bleu}. BLEU is a precision-based metric that considers n-gram overlap between a system output and one or more human reference translations. Despite its usefulness, it is not necessarily ideal to evaluate contextual phenomena as only n-grams of size of to 4 are considered.

Various challenge sets have been constructed to specifically evaluate specific phenomena, such as coreference resolution, pronoun choice and coherence~\citep{bawden2017evaluating, muller2018large}. For this paper, we consider four challenge sets targeting lexical cohesion, deixis (politeness marker), verb phrase ellipsis and morphological inflexion for English$\rightarrow$Russian (En$\rightarrow$Ru).

These four challenge sets consist of multiple-choice questions with identical source sentences, source contexts and target contexts. To answer an example appropriately, the model must assign a higher probability to the correct answer than to the distractors. However, as these are scoring challenges, choosing the right answer does not necessarily entail that the model would generate that exact sentence.

\paragraph {Lexical cohesion} This challenge set evaluates whether a model can consistently translate named entities across sentences. For each example, a named entity in multiple ways, each acceptable without context, but with only one matching a previous translation. For a model to be successful on this task, use of target-side context is beneficial.

While this challenge set only evaluates En$\rightarrow$Ru, maintaining lexical cohesion is a desirable property across many languages. It is split into validation and test subsets, of size 500 and 1500 respectively.

\paragraph {Deixis} While deixis covers multiple sub-phenomena, this challenge set specifically targets the T-V distinction (from the latin $tu$ and $vos$). The T form is informal, while the V form is more polite, used for example when addressing people with higher social standing. The T-V distinction may involve changes to multiple words within the target sentence in order to maintain grammatical correctness.

In addition to Russian, the T-V distinction is also present in other languages such as French (singular \textit{vous}) or Spanish (\textit{t\'u} / \textit{usted}). As for lexical cohesion, use of target-side context is crucial, and there is both validation and test data, of size 500 and 2500.

\paragraph {Verb phrase ellipsis} This challenge set mostly considers the translation of the verb $do$ (including its different conjugations). In Russian and several other languages, it is necessary to specify what is being done. The necessary information may often be found in neighboring sentences, either on the source or target side.

In general, each example within this challenge set contains many more distractors than for lexical cohesion or deixis. In this challenge set only test data is available, so that we can not monitor progress on this phenomenon when training a model.

\paragraph{Ellipsis (inflection)} Russian words are richly inflected. There are two numbers (singular/plural), three genders (masculine/feminine/neuter) and six grammatical cases depending of the usage of a word. In some instances, the grammatical case of a translated word can not be inferred with certainty from the source sentence only, but context may help disambiguate the word. Similarly, in this challenge set only the test data is available.

\subsection{Back-translation}

Back-translation~\citep{sennrich2016improving} is an effective technique to leverage monolingual data, which is generally more plentiful than parallel data, for machine translation. In order to improve a $E \rightarrow F$ model, an inverse direction $F \rightarrow E$ model is trained, generally on parallel data. The latter is used to translate many sentences in the target language $F$ in order to obtain synthetic source sentences. These examples, with a synthetic source and natural target, are then included in the training data alongside the original parallel sentence pairs.

Back-translation can further be improved by different techniques.
The models and training data may be further refined through various rounds of back-translation~\citep{hoang2018iterative}. Adding further noise to the generated source samples may also be beneficial~\citep{edunov2018understanding}. To help the model easily distinguish between the original training data and the back-translated samples, tagged-backtranslation prepends a special token on the source side of back-translated data~\citep{caswell2019tagged} which improves the efficacy of the technique further.

\section{Context completion}

\begin{table*}[t!]
\begin{center}
\begin{tabular}{|c||c|c|}
Sentence & Ru & En \\\hline\hline
Context-3 & {\fontencoding{T2A}\selectfont Крути педали.} & Then pedal faster, Third Wheel. \\\hline
Context-2 & {\fontencoding{T2A}\selectfont Она постоянно выходила из себя.} & She got mad at everything.\\\hline
Context-1 & {\fontencoding{T2A}\selectfont Что мне кажется удивительным.}  & Which I find surprising.. \\\hline
Current & {\fontencoding{T2A}\selectfont Она постоянно выходила из себя.} & She got mad at everything.\\
\end{tabular}
\end{center}
\vspace{-3mm}
\caption{Example augmented with the partial copy heuristic.}
\label{tbl:pc}

\vspace{-4mm}
\end{table*}

Many sources of parallel data include document-level information, such as recent versions of Europarl\footnote{statmt.org/europarl/v9} and News-Commentary\footnote{data.statmt.org/news-commentary/v14}~\citep{barrault-etal-2019-findings}. Nevertheless, there are many datasets where only out-of-context sentence pairs are available~\citep{voita-etal:2019b:ACL}. In this paper, we consider a scenario where some, but not all, parallel data includes document-level information.

We ask how we may fill in the missing contextual information to improve the model quality. In particular, to match evaluation conditions, we modify the training data so that all examples have three source and target sentence pairs as context. Examples for which there already is context are left untouched. We explore three different techniques.

\paragraph {Random context} A simple technique to fill in missing contextual information is to add random sentence pairs. This approach, although not limited to three contextual sentences exclusively, was previously used to create fake documents~\citep{junczysdowmunt:2019:WMT}. As the artificial context will most likely be unrelated to the sentences to translate, this approach may have the adverse effect of biasing models towards ignoring context.

\paragraph {Partial copy heuristic} As an alternative to using random context, we design a heuristic designed to encourage the model to sometimes consider the additional context, as well as to copy words or sub-word tokens.

Given a sentence pair, we build the context by adding a copy of the input, as well as two random sentence pairs. To improve robustness to sentence order, we randomly shuffle the four sentence pairs (2 identical, 2 distinct).

We also tried copying the input multiple times, without any unrelated sentences. However, general performance suffered, most likely because the training and evaluation conditions became too different. An augmented example is included in Table~\ref{tbl:pc}.

\paragraph {Context generation} With access to distinct target-side monolingual data, we extract consecutive segments of four sentences. Given the last sentence, an encoder-decoder model is trained to predict the previous three, similarly to a skip-thought model~\citep{kiros2015skip}. Alternatively, a standard language model could be also be used, although the generation process would have to be modified slightly.

This target-to-context model is applied to the original target side of the parallel data to generate plausible contexts. We observed very repetitive outputs with beam search, so we instead use sampling to recover more diverse contexts.

To obtain the source-side context, we back-translate the sampled context with a reverse-direction model. We also include the original target sentence within the input, but remove the last generated sentence from the output and replace it with the original source. For back-translation, we use a reverse-direction context-aware model trained with the partial copy heuristic.

Especially if the translation system is trained to predict the target context, the potential mismatch between real target sentences and generated samples can potentially affect translation quality. 

\section{Experiments}

\subsection{Parallel data}

Both parallel and monolingual data are sourced from OpenSubtitles2018~\cite{lison2018opensubtitles}, which is a collection of movie and TV show subtitles. We use the same 6 million En$\rightarrow$Ru training examples as in~\citet{voita-etal:2019b:ACL}, out of which 1.5 million have context. For these examples, the previous three source and target sentences are provided. Given the fixed-sized context, many of these training instances have overlapping sentences. We also use the same validation and test sets, for both general performance and BLEU. 

We train a SentencePiece model~\citep{kudo2018sentencepiece} separately on the English and Russian side of the training data, with vocabularies of size 32,768. Given the different tokenization, BLEU scores are not directly comparable to~\citet{voita-etal:2019b:ACL, voita-etal-2019-context}, so we retrain some baseline models. Scores on the lexical cohesion, deixis, ellipsis (VP) and ellipsis (infl.) challenge sets remain comparable.

\subsection{Monolingual data}

We gather document-level Russian monolingual data from OpenSubtitles2018. As the raw data is time-stamped, we concatenate multiple consecutive sentences within the same movie or TV show as long as the time difference between two sentences is at most 2 seconds.

The retrieved strings, which may contain arbitrarily many sentences, are filtered to avoid overlap with all the validation and test sets. More specifically, the last sentence of any validation or test example is prohibited to be within the monolingual training data, but contextual sentences are allowed.\footnote{Whitespace within sentences is removed for filtering to deal with potential tokenization differences.} Strings with less than 4 sentences are removed, while longer segments are split into overlapping 4-sentence examples. The final data contains approximately 26 million examples.

This filtering criterion is only applied to the monolingual data we extract, but not to the parallel data~\citep{voita-etal:2019b:ACL}. In that case, the training and test instances came from different movies, which could still have overlapping sentences.

\subsection{Models}

We use Transformer models~\cite{vaswani2017attention} for all experiments. In particular, with bilingual data only, we use Transformer base models, with 6 encoder and decoder layers, 8-head
attention, model and feed-forward projection dimensions of 512 and 2048 respectively. To reduce overfitting, the dropout rate is set to 20\%.

With additional target-side monolingual data and back-translated source inputs, we additionally train Transformer big models. Even with augmented training data, we observed significant overfitting, and as such set the dropout hyper-parameter to 40\%.

Models are trained on v3 TPUs in a 4x4 topology (32 cores). Baseline models are trained with inputs of size 64 by 128 for each core. Multiple sentences, of maximum length 98, are packed within each row to improve computational efficiency. To allow for longer inputs for context-aware models, batches have size 16 x 512, with maximum length 512. Sentences within each example are concatenated to each other, with a reserved token separating each sentence. All our experiments were performed using the Tensorflow Lingvo \citep{shen2019lingvo} framework.

Model outputs are generated with beam search, with a beam width of 8. BLEU scores are evaluated with an in-house re-implementation of mteval-v13a.\footnote{https://github.com/moses-smt/mosesdecoder/blob/master/scripts/generic/mteval-v13a.pl}

\paragraph {Auxiliary systems} The reverse-direction system used to back-translate additional data is trained on the parallel corpus only, with the missing context replaced with the partial copy heuristic. All the extracted monolingual 4-sentence examples are back-translated, unless their length exceeds 512 subword tokens. Following~\citet{caswell:2019:WMT}, a tag is prepended to the predicted source samples. Note that for the baseline system trained with additional monolingual data, we still use the context-aware Ru$\rightarrow$En model for back-translation, but only add the last sentence pair (out of 4) to the training data.

To train the context generation model, we reformat the monolingual data so that the last sentence within a block of four is used as input to the encoder, while the three previous ones are predicted in a left-to-right manner. To reduce the distance between the input and the first predicted words, we also considered models that generated either sentences or tokens from right-to-left, but settled on the first approach. We use transformer base models to generate the missing context.


\begin{table}[t]
\begin{center}
\begin{tabular}{c||c}
Model & BLEU \\
\hline
Baseline & \textbf{32.3} (31.8) \\
\hline
Concat & 31.0 (30.8) \\
\hline
Random & 31.8 (31.5) \\
\hline
Partial copy & 31.6 (31.3) \\
\hline
Context generation &  31.5 (31.4) \\

\end{tabular}
\end{center}

\vspace{-3mm}
\caption{En$\rightarrow$De BLEU scores, with parallel data only. Validation results in parentheses.}
\label{tbl:bleu}

\vspace{-4mm}
\end{table}

\begin{table*}[t]
\begin{center}
\begin{tabular}{c||c|c|c|c}
Model & Deixis & Lex. cohesion & Ellipsis (infl.) & Ellipsis (VP) \\
\hline
Baseline & 50.0 (50.0) & 45.9 (46.2) & 52.0 & 25.0 \\
\hline
Concat & 83.4 (87.2) & 48.9 (48.2) & 76.0 & 73.8 \\
\hline
Random & 69.9 (68.8) & 45.9 (46.4) & 63.6 & 62.3 \\
\hline
Partial copy & \textbf{86.6} (85.1) & \textbf{74.9} (75.4) & 75.5 & 77.9 \\
\hline
Context generation & 85.6 (86.0) & 60.0 (59.6) & 74.8 & 74.2 \\
\hline
\hline
Concat~\citep{voita-etal:2019b:ACL} & 83.5 & 47.5 & \textbf{76.2} & 76.6 \\
\hline
CADec~\citep{voita-etal:2019b:ACL} & 81.6 & 58.1 & 72.2 & \textbf{80.0} \\

\end{tabular}
\end{center}

\vspace{-3mm}
\caption{En$\rightarrow$De challenge set accuracy, with parallel data only. Validation results in parentheses.}
\label{tbl:challenge}

\vspace{-4mm}
\end{table*}

\begin{table}[t]
\begin{center}
\resizebox{0.48\textwidth}{!}{\begin{tabular}{c|c||c|c}
Model & Run & BLEU & Challenge sets\\
\hline
\multirow{4}{*}{Random} & 1 & 31.9 & 61.5 \\
& 2 & 32.0 & 60.2 \\
& 3 & 31.5 & 60.0 \\
& Avg. & 31.8 $\pm$ 0.2 & 60.6 $\pm$ 0.7 \\
\hline
\hline
\multirow{4}{*}{Partial copy} & 1 & 31.7 & 78.4 \\
& 2 & 31.4 & 78.8 \\
& 3 & 31.6 & 79.0 \\
& Avg. & 31.6 $\pm$ 0.1 & 78.7 $\pm$ 0.2 \\

\end{tabular}}
\end{center}

\vspace{-3mm}
\caption{Robustness experiments across 3 runs with different data and model random seeds. Each challenge set is weighted equally.
}
\label{tbl:robustness}

\vspace{-4mm}
\end{table}

\section{Results}

With parallel data only, En$\rightarrow$Ru BLEU scores are shown in Table~\ref{tbl:bleu}. The concat model, where context is available for only a quarter of the data, trails the baseline by more than 1 BLEU. This model is also prone to severe overfitting as the training and test conditions differ significantly.

By completing the missing context in a variety of ways, the BLEU gap is mostly closed, although the baseline still performs better for En$\rightarrow$Ru. Using the same hyper-parameters, there is still some overfitting, but it is less severe than for the concat model. We note the using random contexts to fill in data is as effective as other methods, at least in terms of BLEU.

In the opposite direction (Ru$\rightarrow$En), the concat model still trails the baseline and overfits quickly, even with added regularization. With the partial copy heuristic, overall performance rises (39.4 BLEU), becoming very similar to the baseline model (39.2 BLEU).\footnote{Given the lack of Ru$\rightarrow$En challenge sets, and as we mainly use Ru$\rightarrow$En for back-translation, we did not train models with all the context completion techniques.} 

On the various challenge sets (Table~\ref{tbl:challenge}), the concat models already shows gains over the context-agnostic baseline, in agreement with previous results reported by ~\citet{voita2018context}. However, the gains for lexical cohesion are minimal.

When missing context is replaced with random sentence pairs, performance drops across all tasks compared to the concat model, but remains above the baseline. As such, the model still benefits from data where real context is available (1.5M examples), but the its ability to capture cross-sentential phenomena is impeded by the additional examples.

With the partial copy heuristic, the model obtains an accuracy of 74.9\% on the lexical cohesion test set, improving by 26\% over the concat model with the original data. Compared to CADec~\cite{voita-etal:2019b:ACL}, a two-stage approach in which a baseline model is trained over the entire parallel corpus, and a post-editing model is built from the subset that has context, there is a gain of 15\%. Over the other challenge sets, performance is comparable or slightly higher (-0.5\% to 4.1\% accuracy difference) than with unmodified training data.

Generating the missing context with separate models, accuracy on the lexical cohesion test set reaches 60.0\%, an improvement over the original data, but still less than with the copying heuristic. Results on the other challenge sets are similar to those obtained with unaugmented parallel data.

\paragraph{Robustness}

To characterize the models' robustness to random fluctuations, we trained three distinct copies for both the random and partial copy context completion heuristics. Both the data generation and model parameter seeds were modified for each run.

Results are reported in Table~\ref{tbl:robustness}. For both data completion techniques, BLEU scores differ by 0.5 or less across runs. Averaging results over the four challenge sets, performance variations between runs with similarly generated contexts remain within 2\%, while there is almost a 20\% difference between the two context imputation approaches.

\begin{table*}[t]
\begin{center}
\begin{tabular}{c||c||c|c|c|c}
Model & BLEU & Deixis & Lex. cohesion & Ellipsis (infl.) & Ellipsis (VP) \\
\hline
Random & \textbf{31.8} & 69.9 & 45.9 & 63.6 & 62.3 \\
\hline
Partial copy - 2 & 31.6 & \textbf{86.6} & 74.9 & 75.5 & 77.9 \\
\hline
Partial copy - 3 & 31.5 & 86.2 & 76.7 & \textbf{76.2} & \textbf{78.2}  \\
\hline
Full copy & 30.93 & 85.4 & \textbf{81.3} & 75.6 & 76.8 \\

\end{tabular}
\end{center}

\vspace{-3mm}
\caption{Progression from random to copied contexts.}
\label{tbl:progression}

\vspace{-4mm}
\end{table*}

\subsection{From random to fully copied contexts}

Using either random or partially copied contexts lead to similar BLEU scores, but very different results on the challenge sets. As such, we are interested in the progression from random contexts to exclusively copying the input when training models. Training examples with existing context (1.5M out of 6M), as well as all test data, remain unchanged.

As the input sentence pair is copied more extensively, BLEU starts to decrease, down to 30.9 when missing context is exclusively replaced by copies of the input sentence pair. This behaviour can likely be attributed to the increasing mismatch between training and inference conditions. Moreover, the model may be over-incentivized to copy the context at the expense of translating the source sentence.

As the input is copied more frequently, performance on the lexical cohesion challenge set improves, reaching 81.3\% accuracy in the most extreme scenario. Nevertheless, as Russian proper nouns are declined, a copying mechanism may not capture all relevant phenomena, especially for words that are not split into multiple tokens.

On each of the other challenge sets, as long as missing context is copied at least once, performance remains within a tight 2\% window. However, as previously remarked, using random contexts leads to much worse results on these test sets.

\subsection{With monolingual data}

With additional monolingual data, which is back-translated in context, and also using larger models, BLEU scores unsurprisingly improve (Table~\ref{tbl:bleu_bt}), reaching scores between 33.0 and 33.4. The context-aware models no longer trail behind the baseline, and the concat model now performs comparably to others. Note that, with bilingual and back-translated data mixed in a ratio close to 1, only approximately $3/8$ of the data has either missing or automatically generated target context, as opposed to $3/4$ with the original bilingual data.

Performance on all challenge sets increases significantly for all models, except for the context-less baseline (Table~\ref{tbl:challenge_bt}). With the original bilingual data and additional back-translated data (concat), accuracy reaches 86.1\% or better on all four test sets. If missing contexts are instead replaced by random ones, performance drops on average by 3.65\%. A model trained with the partial copy heuristic has the best performance on the lexical cohesion test set, at 90.1\%, and other comparable performance to the concat model.

\begin{figure}
\centering
\includegraphics[width=\linewidth]{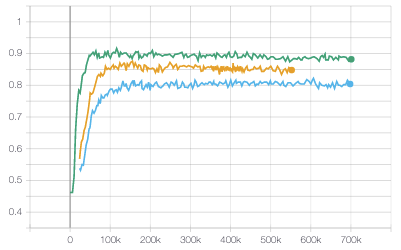}
\caption{Lexical Cohesion validation accuracy over the training process. Orange line corresponds to the Concat model, the blue line corresponds to random context augmentation and the green line represents the partial copy heuristic.}
\label{fig:lex}
\end{figure}

\begin{table*}[t]
\begin{center}
\begin{tabular}{c||c|c|c|c}
Model & Deixis & Lex. cohesion & Ellipsis (infl.) & Ellipsis (VP) \\
\hline
Baseline & 50.0 (50.0) & 45.9 (46.2) & 54.0 & 28.8 \\
\hline
Concat & 87.8 (88.8) & 86.1 (85.6) & \textbf{87.6} & \textbf{88.8} \\
\hline
Random & 85.4 (87.6) & 82.5 (80.0) & 82.8 & 85.2 \\
\hline
Partial copy & 89.6 (89.8) & \textbf{90.1} (89.8) & 86.6 & 88.6 \\
\hline
Context generation & 90.6 (90.8) & 85.3 (83.2) & 85.6 & 87.2 \\
\hline
\hline
DocRepair~\citep{voita-etal-2019-context} & \textbf{91.8} & 80.6 & 86.4 & 75.2 \\\\

\end{tabular}
\end{center}

\vspace{-3mm}
\caption{En$\rightarrow$De challenge set accuracy, with additional back-translated data. Validation results in parentheses.}
\label{tbl:challenge_bt}

\vspace{-4mm}
\end{table*}

Figure~\ref{fig:lex} displays the progress of lexical cohesion validation accuracy for a few models. The model trained with the copy heuristic, which achieves the overall best performance, also learns the fastest. If we instead use random contexts, final performance remains within 10\%, while learning is much slower.

We also present results for the DocRepair model~\cite{voita-etal-2019-context}, where the outputs of a sentence-to-sentence baseline are refined in context. The post-editing model is trained from round-trip translated monolingual data (without context), so that the output is contextually coherent, but not necessarily the input. Note that while we use monolingual data from the same corpus, and a similar amount of examples, our filtering procedures differ, so the training data is not exactly the same.

DocRepair obtains the best performance on the deixis test set, with a margin of 1.2\% over our best model on this task. Results on the ellipsis (infl.) challenge sets are similar to those obtained with single-pass context-aware translation systems. However, these single-pass models perform up to 9.5\% and 13.6\% better on the lexical cohesion and VP ellipsis challenge sets respectively. In particular, while VP ellipsis may be a hard phenomena to capture with monolingual data only~\cite{voita-etal-2019-context}, a context-aware translation system trained on sufficiently many examples may perform well on this task.

\begin{table}[t]
\begin{center}
\begin{tabular}{c||c}
Model & BLEU \\
\hline
Baseline & 33.0 (33.1) \\
\hline
Concat & 33.2 (33.1) \\
\hline
Random & \textbf{33.4} (33.3) \\
\hline
Partial copy & \textbf{33.4} (33.1) \\
\hline
Context generation & \textbf{33.4} (33.2) \\

\end{tabular}
\end{center}

\vspace{-3mm}
\caption{En$\rightarrow$De BLEU scores, with additional back-translated data. Validation results in parentheses.}
\label{tbl:bleu_bt}

\vspace{-4mm}
\end{table}

\section{Conclusion}

When document-level context is only partially available, we evaluate the effectiveness of various data completion techniques, using both BLEU and challenge sets targeting specific linguistic phenomena. In particular, a simple copy heuristic helps models achieve much better lexical cohesion, even for a highly inflected language such as Russian. Only adding random context sentence pairs however reduces a model's ability to capture cross-sentence interactions, yet these effects are not visible from BLEU scores. Additionally, we confirm the effectiveness of back-translation on overall translation quality, while also demonstrating its usefulness on the four challenge sets.

As even simple context completion techniques have a clear impact on model performance, it may be worthwhile to explore additional approaches. In particular, to obtain more natural contexts, while limiting model generation errors, we can envision embedding sentences such that neighbours in vector space are probable contexts of each other. Moreover, it might be useful to understand how different data augmentation schemes interact with more sophisticated model architectures.

\section*{Acknowledgments}
We would like to thank Kyunghyun Cho for insightful comments and discussions, the Google Translate and Google Brain teams for their useful inputs. \\

\bibliography{emnlp-ijcnlp-2019}

\begin{thebibliography}{33}
\expandafter\ifx\csname natexlab\endcsname\relax\def\natexlab#1{#1}\fi

\bibitem[{Bahdanau et~al.(2015)Bahdanau, Cho, and Bengio}]{bahdanau2014neural}
Dzmitry Bahdanau, Kyunghyun Cho, and Yoshua Bengio. 2015.
\newblock Neural machine translation by jointly learning to align and
  translate.
\newblock In \emph{International Conference on Learning Representations
  (ICLR)}.

\bibitem[{Barrault et~al.(2019)Barrault, Bojar, Costa-juss{\`a}, Federmann,
  Fishel, Graham, Haddow, Huck, Koehn, Malmasi, Monz, M{\"u}ller, Pal, Post,
  and Zampieri}]{barrault-etal-2019-findings}
Lo{\"\i}c Barrault, Ond{\v{r}}ej Bojar, Marta~R. Costa-juss{\`a}, Christian
  Federmann, Mark Fishel, Yvette Graham, Barry Haddow, Matthias Huck, Philipp
  Koehn, Shervin Malmasi, Christof Monz, Mathias M{\"u}ller, Santanu Pal, Matt
  Post, and Marcos Zampieri. 2019.
\newblock \href {https://www.aclweb.org/anthology/W19-5301} {Findings of the
  2019 conference on machine translation ({WMT}19)}.
\newblock In \emph{Proceedings of the Fourth Conference on Machine Translation
  (Volume 2: Shared Task Papers, Day 1)}, pages 1--61, Florence, Italy.
  Association for Computational Linguistics.

\bibitem[{Bawden et~al.(2018)Bawden, Sennrich, Birch, and
  Haddow}]{bawden2017evaluating}
Rachel Bawden, Rico Sennrich, Alexandra Birch, and Barry Haddow. 2018.
\newblock Evaluating discourse phenomena in neural machine translation.
\newblock In \emph{Proceedings of the 2018 Conference of the North American
  Chapter of the Association for Computational Linguistics: Human Language
  Technologies, Volume 1 (Long Papers)}, volume~1, pages 1304--1313.

\bibitem[{Caswell et~al.(2019{\natexlab{a}})Caswell, Chelba, and
  Grangier}]{caswell:2019:WMT}
Isaac Caswell, Ciprian Chelba, and David Grangier. 2019{\natexlab{a}}.
\newblock Tagged back-translation.
\newblock In \emph{Proceedings of the Fourth Conference on Machine
  Translation}.

\bibitem[{Caswell et~al.(2019{\natexlab{b}})Caswell, Chelba, and
  Grangier}]{caswell2019tagged}
Isaac Caswell, Ciprian Chelba, and David Grangier. 2019{\natexlab{b}}.
\newblock Tagged back-translation.
\newblock \emph{arXiv preprint arXiv:1906.06442}.

\bibitem[{Edunov et~al.(2018)Edunov, Ott, Auli, and
  Grangier}]{edunov2018understanding}
Sergey Edunov, Myle Ott, Michael Auli, and David Grangier. 2018.
\newblock Understanding back-translation at scale.
\newblock In \emph{Proceedings of the 2018 Conference on Empirical Methods in
  Natural Language Processing}, pages 489--500.

\bibitem[{Gehring et~al.(2017)Gehring, Auli, Grangier, Yarats, and
  Dauphin}]{gehring2017convolutional}
Jonas Gehring, Michael Auli, David Grangier, Denis Yarats, and Yann~N Dauphin.
  2017.
\newblock Convolutional sequence to sequence learning.
\newblock In \emph{International Conference on Machine Learning}, pages
  1243--1252.

\bibitem[{Hassan et~al.(2018)Hassan, Aue, Chen, Chowdhary, Clark, Federmann,
  Huang, Junczys-Dowmunt, Lewis, Li et~al.}]{hassan2018achieving}
Hany Hassan, Anthony Aue, Chang Chen, Vishal Chowdhary, Jonathan Clark,
  Christian Federmann, Xuedong Huang, Marcin Junczys-Dowmunt, William Lewis,
  Mu~Li, et~al. 2018.
\newblock Achieving human parity on automatic chinese to english news
  translation.
\newblock \emph{arXiv preprint arXiv:1803.05567}.

\bibitem[{He et~al.(2018)He, Tan, Xia, He, Qin, Chen, and Liu}]{he2018layer}
Tianyu He, Xu~Tan, Yingce Xia, Di~He, Tao Qin, Zhibo Chen, and Tie-Yan Liu.
  2018.
\newblock \href
  {http://papers.nips.cc/paper/8019-layer-wise-coordination-between-encoder-and-decoder-for-neural-machine-translation.pdf}
  {Layer-wise coordination between encoder and decoder for neural machine
  translation}.
\newblock In S.~Bengio, H.~Wallach, H.~Larochelle, K.~Grauman, N.~Cesa-Bianchi,
  and R.~Garnett, editors, \emph{Advances in Neural Information Processing
  Systems 31}, pages 7944--7954. Curran Associates, Inc.

\bibitem[{Hoang et~al.(2018)Hoang, Koehn, Haffari, and
  Cohn}]{hoang2018iterative}
Vu~Cong~Duy Hoang, Philipp Koehn, Gholamreza Haffari, and Trevor Cohn. 2018.
\newblock Iterative back-translation for neural machine translation.
\newblock In \emph{Proceedings of the 2nd Workshop on Neural Machine
  Translation and Generation}, pages 18--24.

\bibitem[{Jean et~al.(2017)Jean, Lauly, Firat, and Cho}]{jean2017does}
Sebastien Jean, Stanislas Lauly, Orhan Firat, and Kyunghyun Cho. 2017.
\newblock Does neural machine translation benefit from larger context?
\newblock \emph{arXiv preprint arXiv:1704.05135}.

\bibitem[{Junczys-Dowmunt(2019)}]{junczysdowmunt:2019:WMT}
Marcin Junczys-Dowmunt. 2019.
\newblock \href {http://www.aclweb.org/anthology/W19-5321} {Microsoft
  translator at wmt 2019: Towards large-scale document-level neural machine
  translation}.
\newblock In \emph{Proceedings of the Fourth Conference on Machine Translation
  (Volume 2: Shared Task Papers, Day 1)}, pages 225--233, Florence, Italy.
  Association for Computational Linguistics.

\bibitem[{Kiros et~al.(2015)Kiros, Zhu, Salakhutdinov, Zemel, Urtasun,
  Torralba, and Fidler}]{kiros2015skip}
Ryan Kiros, Yukun Zhu, Ruslan~R Salakhutdinov, Richard Zemel, Raquel Urtasun,
  Antonio Torralba, and Sanja Fidler. 2015.
\newblock Skip-thought vectors.
\newblock In \emph{Advances in neural information processing systems}, pages
  3294--3302.

\bibitem[{Kudo and Richardson(2018)}]{kudo2018sentencepiece}
Taku Kudo and John Richardson. 2018.
\newblock Sentencepiece: A simple and language independent subword tokenizer
  and detokenizer for neural text processing.
\newblock In \emph{Proceedings of the 2018 Conference on Empirical Methods in
  Natural Language Processing: System Demonstrations}, pages 66--71.

\bibitem[{L{\"a}ubli et~al.(2018)L{\"a}ubli, Sennrich, and
  Volk}]{laubli2018has}
Samuel L{\"a}ubli, Rico Sennrich, and Martin Volk. 2018.
\newblock Has machine translation achieved human parity? a case for
  document-level evaluation.
\newblock In \emph{Proceedings of the 2018 Conference on Empirical Methods in
  Natural Language Processing}, pages 4791--4796.

\bibitem[{Lison et~al.(2018)Lison, Tiedemann, and
  Kouylekov}]{lison2018opensubtitles}
Pierre Lison, J{\"o}rg Tiedemann, and Milen Kouylekov. 2018.
\newblock Opensubtitles2018: Statistical rescoring of sentence alignments in
  large, noisy parallel corpora.
\newblock In \emph{Proceedings of the Eleventh International Conference on
  Language Resources and Evaluation (LREC-2018)}.

\bibitem[{Mansimov et~al.(2019)Mansimov, Wang, and
  Cho}]{mansimov2019generalized}
Elman Mansimov, Alex Wang, and Kyunghyun Cho. 2019.
\newblock A generalized framework of sequence generation with application to
  undirected sequence models.
\newblock \emph{arXiv preprint arXiv:1905.12790}.

\bibitem[{Maruf and Haffari(2018)}]{maruf2017document}
Sameen Maruf and Gholamreza Haffari. 2018.
\newblock Document context neural machine translation with memory networks.
\newblock In \emph{Proceedings of the 56th Annual Meeting of the Association
  for Computational Linguistics (Volume 1: Long Papers)}, volume~1, pages
  1275--1284.

\bibitem[{Miculicich et~al.(2018)Miculicich, Ram, Pappas, and
  Henderson}]{miculicich2018document}
Lesly Miculicich, Dhananjay Ram, Nikolaos Pappas, and James Henderson. 2018.
\newblock Document-level neural machine translation with hierarchical attention
  networks.
\newblock In \emph{Proceedings of the 2018 Conference on Empirical Methods in
  Natural Language Processing}, pages 2947--2954.

\bibitem[{M{\"u}ller et~al.(2018)M{\"u}ller, Rios, Voita, and
  Sennrich}]{muller2018large}
Mathias M{\"u}ller, Annette Rios, Elena Voita, and Rico Sennrich. 2018.
\newblock A large-scale test set for the evaluation of context-aware pronoun
  translation in neural machine translation.
\newblock \emph{arXiv preprint arXiv:1810.02268}.

\bibitem[{Papineni et~al.(2002)Papineni, Roukos, Ward, and
  Zhu}]{papineni2002bleu}
Kishore Papineni, Salim Roukos, Todd Ward, and Wei-Jing Zhu. 2002.
\newblock Bleu: a method for automatic evaluation of machine translation.
\newblock In \emph{Proceedings of the 40th annual meeting on association for
  computational linguistics}, pages 311--318. Association for Computational
  Linguistics.

\bibitem[{Popescu-Belis(2019)}]{popescu2019context}
Andrei Popescu-Belis. 2019.
\newblock Context in neural machine translation: A review of models and
  evaluations.
\newblock \emph{arXiv preprint arXiv:1901.09115}.

\bibitem[{Sennrich et~al.(2016)Sennrich, Haddow, and
  Birch}]{sennrich2016improving}
Rico Sennrich, Barry Haddow, and Alexandra Birch. 2016.
\newblock Improving neural machine translation models with monolingual data.
\newblock In \emph{Proceedings of the 54th Annual Meeting of the Association
  for Computational Linguistics (Volume 1: Long Papers)}, pages 86--96.

\bibitem[{Shen et~al.(2019)Shen, Nguyen, Wu, Chen et~al.}]{shen2019lingvo}
Jonathan Shen, Patrick Nguyen, Yonghui Wu, Zhifeng Chen, et~al. 2019.
\newblock \href {http://arxiv.org/abs/1902.08295} {Lingvo: a modular and
  scalable framework for sequence-to-sequence modeling}.

\bibitem[{Sutskever et~al.(2014)Sutskever, Vinyals, and
  Le}]{sutskever2014sequence}
Ilya Sutskever, Oriol Vinyals, and Quoc~V Le. 2014.
\newblock Sequence to sequence learning with neural networks.
\newblock \emph{NIPS}.

\bibitem[{Tiedemann and Scherrer(2017)}]{tiedemann2017neural}
J{\"o}rg Tiedemann and Yves Scherrer. 2017.
\newblock Neural machine translation with extended context.
\newblock In \emph{Proceedings of the Third Workshop on Discourse in Machine
  Translation}, pages 82--92.

\bibitem[{Tu et~al.(2018)Tu, Liu, Shi, and Zhang}]{tu2018learning}
Zhaopeng Tu, Yang Liu, Shuming Shi, and Tong Zhang. 2018.
\newblock Learning to remember translation history with a continuous cache.
\newblock \emph{Transactions of the Association of Computational Linguistics},
  6:407--420.

\bibitem[{Vaswani et~al.(2017)Vaswani, Shazeer, Parmar, Uszkoreit, Jones,
  Gomez, Kaiser, and Polosukhin}]{vaswani2017attention}
Ashish Vaswani, Noam Shazeer, Niki Parmar, Jakob Uszkoreit, Llion Jones,
  Aidan~N Gomez, {\L}ukasz Kaiser, and Illia Polosukhin. 2017.
\newblock Attention is all you need.
\newblock In \emph{Advances in Neural Information Processing Systems}, pages
  5998--6008.

\bibitem[{Voita et~al.(2019{\natexlab{a}})Voita, Sennrich, and
  Titov}]{voita-etal-2019-context}
Elena Voita, Rico Sennrich, and Ivan Titov. 2019{\natexlab{a}}.
\newblock Context-aware monolingual repair for neural machine translation.
\newblock In \emph{Proceedings of the 2019 Conference on Empirical Methods in
  Natural Language Processing and 9th International Joint Conference on Natural
  Language Processing}, Hong Kong, China. Association for Computational
  Linguistics.

\bibitem[{Voita et~al.(2019{\natexlab{b}})Voita, Sennrich, and
  Titov}]{voita-etal:2019b:ACL}
Elena Voita, Rico Sennrich, and Ivan Titov. 2019{\natexlab{b}}.
\newblock \href {https://arxiv.org/pdf/1905.05979.pdf} {{When a Good
  Translation is Wrong in Context: Context-Aware Machine Translation Improves
  on Deixis, Ellipsis, and Lexical Cohesion}}.
\newblock In \emph{{Proceedings of the 57th Annual Meeting of the Association
  for Computational Linguistics}}, Florence, Italy. Association for
  Computational Linguistics.

\bibitem[{Voita et~al.(2018)Voita, Serdyukov, Sennrich, and
  Titov}]{voita2018context}
Elena Voita, Pavel Serdyukov, Rico Sennrich, and Ivan Titov. 2018.
\newblock Context-aware neural machine translation learns anaphora resolution.
\newblock In \emph{Proceedings of the 56th Annual Meeting of the Association
  for Computational Linguistics (Volume 1: Long Papers)}, volume~1, pages
  1264--1274.

\bibitem[{Wang et~al.(2017)Wang, Tu, Way, and Liu}]{wang2017exploiting}
Longyue Wang, Zhaopeng Tu, Andy Way, and Qun Liu. 2017.
\newblock Exploiting cross-sentence context for neural machine translation.
\newblock In \emph{Proceedings of the 2017 Conference on Empirical Methods in
  Natural Language Processing}, pages 2826--2831.

\bibitem[{Zhang et~al.(2018)Zhang, Luan, Sun, Zhai, Xu, Zhang, and
  Liu}]{zhang2018improving}
Jiacheng Zhang, Huanbo Luan, Maosong Sun, Feifei Zhai, Jingfang Xu, Min Zhang,
  and Yang Liu. 2018.
\newblock Improving the transformer translation model with document-level
  context.
\newblock In \emph{Proceedings of the 2018 Conference on Empirical Methods in
  Natural Language Processing}, pages 533--542.

\end{thebibliography}
\bibliographystyle{acl_natbib}

\end{document}